%% file: dap_2018_arxiv_acm.tex
\documentclass[sigconf, authorversion]{acmart}

\usepackage{booktabs} 
\usepackage{enumerate}

\DeclareMathOperator*{\argmin}{arg\,min}

\newcommand{\beq}{\begin{equation}}
\newcommand{\eeq}{\end{equation}}

\newcommand{\z}{\mathbf{z}}
\newcommand{\x}{\mathbf{x}}
\newcommand{\w}{\mathbf{w}}
\newcommand{\E}{\mathbf{E}}
\DeclareMathOperator{\Tr}{Tr}
\DeclareMathOperator{\diag}{diag}

\setcopyright{none}

\acmConference[AAAI-18]{Thirty-Second AAAI Conference on Artificial Intelligence}{February 2018}{New Orleans, Louisiana, USA}
\acmYear{2018}
\copyrightyear{2018}
\acmArticle{}
\acmPrice{}
\acmDOI{}
\acmISBN{}

\begin{document}
\title{Topic Modeling on Health Journals with Regularized Variational Inference}

\author{Robert Giaquinto}
\affiliation{%
  \institution{University of Minnesota}
  \streetaddress{Keller Hall, 200 Union Street SE}
  \city{Minneapolis}
  \state{MN}
  \postcode{55455}
}
\email{giaquinto.ra@gmail.com}

\author{Arindam Banerjee}
\affiliation{%
  \institution{University of Minnesota}
  \streetaddress{Keller Hall, 200 Union Street SE}
  \city{Minneapolis}
  \state{MN}
  \postcode{55455}
}
\email{banerjee42@gmail.com}

\begin{abstract}
Topic modeling enables exploration and compact representation of a corpus. The CaringBridge (CB) dataset is a massive collection of journals written by patients and caregivers during a health crisis. Topic modeling on the CB dataset, however, is challenging due to the asynchronous nature of multiple authors writing about their health journeys. To overcome this challenge we introduce the Dynamic Author-Persona topic model (DAP), a probabilistic graphical model designed for temporal corpora with multiple authors. The novelty of the DAP model lies in its representation of authors by a persona --- where personas capture the propensity to write about certain topics over time. Further, we present a regularized variational inference algorithm, which we use to encourage the DAP model's personas to be distinct. Our results show significant improvements over competing topic models --- particularly after regularization, and highlight the DAP model's unique ability to capture common journeys shared by different authors.
\end{abstract}

\keywords{machine learning, topic modeling, graphical model, regularized variational inference, healthcare}

\maketitle

\input{1_introduction}

\input{2_background}

\input{3_model}

\input{4_variational_em}

\input{5_experiments}

\input{6_results}

\input{7_conclusion}

\input{8_acknowledgements}

\bibliographystyle{ACM-Reference-Format}
\bibliography{dap_2018_arxiv_acm}

\input{9_appendix}

\end{document}

%% file: 1_introduction.tex
\section{Introduction}
Topic models can compactly represent large collections of documents by the themes running through them. We introduce a topic model designed for the unique challenges presented by the CaringBridge (CB) dataset. The CB dataset includes journals written by patients and caregivers during a health crisis. CB journals function like a blog, and are shared to a private community of friends and family. The full dataset includes 13.1 million journals written by approximately half a million authors between 2006 and 2016. From the CB dataset we're interested in capturing health journeys, that is, authors writing about the same topics over time.

The challenges in topic modeling on the CB dataset stem from the asynchronous nature of author's posts. Specifically, authors start and stop journaling at different times --- both in terms of calendar dates and how far along they are in their health journey. Additionally, authors post at irregular frequencies. While about 15\% of CB authors post nearly everyday, the majority of authors typically post less frequently, often corresponding to a major update, event, or anniversary of an event. What's more, the length of these posts can range from just a few words to thousands of words.

State-of-the-art topic models can identify topics \cite{Blei2003}, track how topics change over time \cite{Blei2006a,Wang2006a,Wei2007,Wang2008}, or associate authors with certain topics \cite{Rosen-Zvi2004,Steyvers2004a,McCallum2005,Mimno2007}. These models cannot, however, describe common narratives and the author's sharing them. We present the Dynamic Author-Persona topic model (DAP), a novel approach that represents authors by latent \emph{personas}. Personas act as a soft-clustering on authors based their propensity to write about similar topics over time. Our approach is unique in multiple respects. First, unlike other temporal topic models, the words making up a topic don't evolve over time --- rather, DAP's personas reflect the flow of conversation from one topic to next. Second, we introduce a regularized variational inference (RVI) algorithm, an approach we use to encourage personas to be distinct from one another.

Our results show that the DAP model outperforms competing topic models, producing better likelihoods on heldout data. Finally, we demonstrate that using RVI further improves the DAP model's performance, and results in personas that are rich and compelling descriptions of the health journeys experienced by CB authors.

The rest of the paper is as follows: in Section \ref{background}, a brief background on temporal topic models is given. Section \ref{model} presents the DAP model. Section \ref{inference} details the model's RVI algorithm. Section \ref{experiments} introduces the evaluation dataset and procedure. Section \ref{results} shares the results of the experiments. Finally, in Section \ref{conclusion} we summarize the contributions of this paper.

%% file: 2_background.tex
\section{Background}
\label{background}
Much of the research on topic modeling builds on the latent Dirichlet allocation (LDA) model \cite{Blei2003}. The LDA model doesn't account for meta-information like authorship or time. Nevertheless, interest in LDA has endured, in part, due its ability to richly describe topics as distributions over words and documents as mixtures of topics. In the years since LDA's introduction, others have extended the idea to compliment corpora with a variety of structures and metadata.

Author information is common in many corpora. A few topic models are designed to identify authors' preferences for certain topics, and the relationships between authors \cite{Rosen-Zvi2004,Steyvers2004a,McCallum2005,Mimno2007,Pathak2008}.  Corpora with a temporal structure, such as scientific journals or newspaper articles, are the focus of many of temporal topic models \cite{Blei2006a,Wang2006a,Wei2007,Wang2008}.

{\bf Temporal Topic Models.} Two topic models set the standard of comparison for topic modeling on corpora with a temporal element: the dynamic topic model (DTM) \cite{Blei2006a} and the topics over time model TOT \cite{Wang2006a}. These two models represent very different approaches to modeling time in a topic model.

The TOT model defines time as an observed variable, which leads to a continuous treatment of time and the ability to predict timestamps of documents. Alternatively, the DTM evolves topics over time using a Markov process. In many corpora the evolution of topics provides interesting insights. For example, Blei's model of the \emph{Science} corpus shows words associated with a topic on physics changing over a century.

Building directly on the DTM, in 2008 Wang et al. developed the continuous time dynamic topic model (CDTM) which uses continuous Brownian motion to model the evolution of topics over time \cite{Wang2008}. This is a major development in temporal topic models because, unlike the DTM, it doesn't require partitioning the data into discrete time periods. Instead, the model assumes that at each time step the variance in the topic proportions increases proportional to the duration since the previous document. Similar to Wang et al., the Dynamic Mixture Model (DMM) is built for continuous streams of text \cite{Wei2007}. In the DMM, however, topics are fixed in time and the model captures the evolution of document-level topic proportions over time.

{\bf Topic Modeling of Health Journeys.} In many topic modeling applications to temporal corpora, the time component is ignored. For example, Wen et al. model cancer event trajectories from users of an online forum for breast cancer support \cite{Wen2012}. Wen's approach uses LDA to extract cancer event keywords, which are then linked together in time by temporal descriptions mined from the text. This work demonstrates a quantitative approach to studying the dynamics of social support network, and offer a powerful look at the experiences of users in these support networks.

Numerous studies have shown that support networks, both in person and online, are valuable tools for those suffering from chronic conditions or life-threating illness and caregivers \cite{Wen2011,Rodgers2005,Beaudoin2007}. Additionally, online social networks can serve as a way to efficiently disseminate information regarding someone's status to their community. Understanding the health journeys of users in these social support communities is valuable information for improving user experience. Topic models are uniquely suited to succinctly describing and analyzing these health journeys.

%% file: 3_model.tex
\section{The DAP Model}
\label{model}

The design of the DAP model was made with journaling behavior in mind. Consider a CB author journaling about their surgery: initially they may write about topics related to the surgical procedure, but as time progresses the author is more likely to discuss recovery, physical therapy, or returning to normal life. In other words, the likelihood of a topic for some document depends where the document's author is in their health journey. As such, DAP assumes that (1) a state space model controls the likelihood of a topic at each time step, (2) each persona represents a different flow of topics over time, and (3) each author has a distribution over personas.

The DAP's approach for modeling topics in a document, and words in a topic follows the correlated topic model (CTM) and LDA, respectively \cite{Lafferty2006,Blei2003}. The idea of modeling latent personas was originally proposed by \cite{Mimno2007}, however in their Author-Persona Topic model (APT) personas differ significantly from those proposed in the DAP model. First, in the APT model each author may have a different number of personas, whereas DAP models each author as a distribution over a fixed number of personas --- which acts as a soft clustering on authors. Second, APT assigns a persona to documents, which indirectly defines a topic distribution for each cluster of documents, whereas we model documents as each having their own distribution over topics. Lastly, while DAP's personas also correspond to a distribution over topics, DAP evolves these topic distributions over time --- thereby capturing the inherent temporal structure resulting from an author writing multiple documents.

The DAP model directly addresses the challenges presented by the CB dataset. First, the asynchronous nature of health journals is handled by: (1) transforming each journal's timestamp to the time elapsed since the author's first post, and (2) learning multiple personas to account for a wide variety in topic trajectories. Second, irregular posting behavior is managed by employing the Brownian motion model, originally used in topic modeling by \cite{Wang2008}, to model topic variance as proportional to the gap in time between documents.

The generative process of the model is described below. The model assumes that each document $d$ in the corpus has a timestamp $s_t$ associated with it. Similar to the CDTM \cite{Wang2008}, timestamps are used in a continuous Brownian motion model to capture an increase in topic variance as time between observations increases. More formally, if $s_i$ and $s_j$ are timestamps at steps $j > i > 0$, then $\Delta_{s_j, s_i}$ is the difference in time between $s_j$ and $s_i$. We use the shorthand $\Delta_{s_t}$ to denote the difference in time between timestamps $s_t$ and $s_{t-1}$. For brevity the variance $\sigma \Delta_{s_t} I$ is denoted $\Sigma_t$, where $\sigma$ is a known process noise in the state space model.

\begin{enumerate}
\item Draw distribution over words $\beta_k \sim Dir(\eta)$ for each topic $k$.
\item Draw distribution over personas $\kappa_a \sim Dir(\omega)$ for each author $a$.
\item For each persona $p$, draw initial distribution over topics:

\begin{equation*}
\alpha_{0,p} \sim \mathcal{N}(\mu_0, \Sigma_0), \forall p \in \{1, \dots, P \}~.
\end{equation*}

\item For each time step $t$, where $t \in \{1, \dots, T\}$:
    \begin{itemize}
    \item Draw distribution over topics:
    \begin{equation*}
    \alpha_{t,p} \sim \mathcal{N}(\alpha_{t-1,p}, \Sigma_{t-1}), \forall p \in \{1, \dots, P\}~.
    \end{equation*}
    \item Update $\Sigma_t$ according to Brownian motion model: $\Sigma_t - \Sigma_{t-1} \sim \mathcal{N}(0, \sigma \Delta_{s_t} I)$.
    \item For each document $d$, where $d \in \{1, \dots, D_t\}$:
    	\begin{enumerate}[(a)]
    	\item Choose persona indicator $x_{t,d} \sim Mult(\kappa_a)$ where $a$ corresponds to the author of document $d_t$.
    	\item Draw topic distribution $\theta_{t,d} \sim \mathcal{N}(\alpha_t x_{t,d}, \Sigma_t)$ for document $d_t$.
    	\item For each word $w_{t,d,n}$, where $n \in \{1, \dots, N_{d_t} \}$:
    		\begin{enumerate}[i.]
    		\item Choose word topic indicator $z_n \sim Mult(\pi(\theta_{t,d}))$.
    		\item Choose word $w_{t,d,n}$ from $p(w_{t,d,n} \mid \beta_{z_n})$, a multinomial probability conditioned on the topic indicator $z_n$.
    		\end{enumerate}
    	\end{enumerate}
    \end{itemize}
\end{enumerate}

Following the approach in the CTM and DTM, we use the function $\pi(\cdot)$ to map the Logistic Normal $\theta_{t,d}$, parameterized by a mean $\alpha_{t,k,p}$ and covariance $\sigma \Delta_{s_t} I$, to the multinomial's natural parameters via $\pi(\theta_{t,d}) = \frac{\exp(\theta_{t,d})}{\sum_d^D \exp(\theta_{t,d})}$ in order to obey the constraint that the parameters lie on the simplex.

\begin{figure}
\centering
\includegraphics[width=1.0\linewidth]{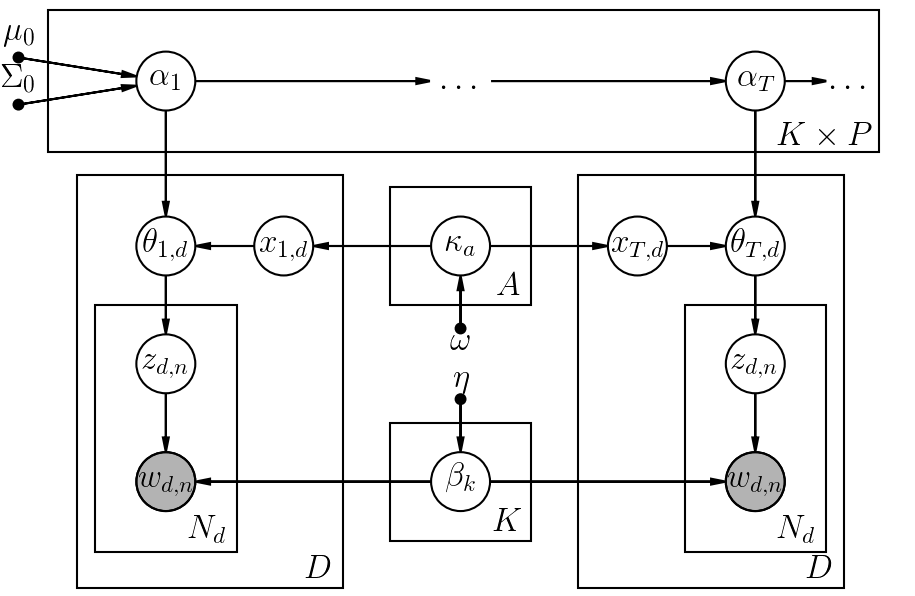}
\caption{Graphical representation of the Dynamic Author-Persona topic model (DAP). On top, topic distributions for each persona evolve over time: $\alpha_t | \alpha_{t-1} \sim \mathcal{N}(\alpha_{t-1}, \Sigma)$. The distribution over words for each topic, $\beta \sim Dir(\eta)$, is fixed in time. Each author $a \in \{1, \dots, A\}$ is represented by a distribution over personas, that is $\kappa_a \sim Dir(\omega)$. The distribution over topics for each document is dependent on the persona distribution $\x_{t,d}$ for that document's author, and the evolving topic distribution $\alpha_{t}$.}
\label{fig:dap}
\end{figure}

The graphical model corresponding to this process is shown in Figure \ref{fig:dap}. In LDA and its extensions the parameter $\alpha$ represents a prior probability of each topic. In the DAP model, $\alpha_{t,1:K,p}$ takes on an expanded role: it's a distribution over $K$ topics at time step $t$ for persona $p$. The choice of letting $\alpha$ evolve over time, as opposed to $\beta$ like in the DTM, is that in a collection of journals there is less interest in changes to topics themselves. In other words, we model the words associated with a topic as static in time, but the topics an author writes about will change over time.

%% file: 4_variational_em.tex
\section{Variational EM Algorithm}
\label{inference}

Given the model structure, next we derive an inference algorithm used to estimate the model's latent parameters. Much like LDA and its extensions, the DAP model's posterior:

\begin{equation*}
p(\kappa, \x, \alpha, \beta, \theta, \z \mid \w, \omega, \eta) = \frac{p(\kappa, \x, \alpha, \beta, \theta, \z, \w \mid \omega, \eta)}{p(\w \mid \mu_0, \sigma_0, \eta, \omega)}~,
\end{equation*}

\noindent is intractable due to the normalization term. In order learn optimal values to the model's parameters we use a form of variational inference (VI), which approximates the difficult to compute posterior distribution $p$ with a simpler distribution $q$ (see \citeauthor{Blei2016}, 2016 for a review). Variational inference casts an inference problem as an optimization problem with the goal of finding parameters to the variational distribution such that $q = q(\kappa, \x, \alpha, \theta, \z, \beta)$ closely approximates $p=p(\kappa, \x, \alpha, \theta, \z, \beta \mid \w)$. Our regularized variational inference (RVI) algorithm seeks a distribution $q \in \mathcal{Q}$ such that

\begin{equation}
\label{eqn:kl}
q^* = \argmin_{q \in \mathcal{Q}} KL( q \mid \mid p) + \rho r(q)~,
\end{equation}

\noindent where $KL(\cdot)$ is KL-Divergence. The added term $r(q)$ is a regularization function we've introduced to discourage similar personas (further detail given in Section \ref{sec:penalty}), and $\rho$ the corresponding hyperparameter.

To make $q$ easy to compute, we apply \emph{mean field} variational inference which assumes that the parameters are \emph{posteriori} independent. Under the mean field assumption the variational distribution factorizes as:

\begin{equation}
\begin{split}
\prod_{k=1}^K	& q(\beta_k \mid \lambda_k) \prod_{a=1}^A  q(\kappa_a \mid \delta_a) \prod_{p=1}^P q(\alpha_{1:T,k,p} \mid \hat{\alpha}_{1:T,k,p}) \times\\
& \prod_{t=1}^T \prod_{d=1}^{D_t} q(x_{t,d,p} \mid \tau_{t,d,p}) q(\theta_{t,d} \mid \gamma_{t,d}) \prod_{n=1}^{N_{d_t}} q(z_n \mid \phi_n)
\end{split}
\end{equation}

\noindent where we have introduced the following variational parameters: the persona for each author $\kappa_a$ is endowed with a free Dirichlet parameter $\delta_a$; each assignment of a persona to an author $\x_{t,d}$ is endowed with a free Multinomial parameter $\tau_{t,d}$; in the variational distribution of $\alpha_{1:T,k,p}$ the sequential structure is kept intact with variational observations $\hat{\alpha}_{1:T,k,p}$; each document-topic proportion vector $\theta_{t,d}$ is endowed with a free $\gamma_d$. The variance for the document-topic parameters are $v_{t,d}$ and $\hat{v}_{t,d}$, for the model and variational parameter, respectively; each word-topic indicator is endowed with a free multinomial parameter $\phi_{t,d,n}$.

An optimal $q$ cannot be computed directly, but following \citeauthor{Jordan1999} (1999) an optimization of the variational parameters proceeds by maximizing a bound on the log-likelihood of the data. In the DAP model, data is observed in words $\w$ for each document $d$ at time step $t$, hence we re-write the log-likelihood of the data $\log p(\w_{t,d})$ by:

\begin{equation}
\begin{split}
\label{eqn:lhood}
\log p(\w_{t,d}) &= \log \iiiint \sum_{\z} \sum_{\x} p \hspace{.2cm} \partial \theta_{t,d}, \partial \alpha_t, \partial \beta, \partial \kappa \\
&= \log \iiiint \sum_{\z} \sum_{\x} \frac{p q}{q} \\
& \geq \iiiint \sum_{\z} \sum_{\x} q \log p - \iiiint \sum_{\z} \sum_{\x} q \log q \\
&= \E_q[ \log p] - \E_q[\log q]
\end{split}
\end{equation}

The inequality in \eqref{eqn:lhood} follows from Jensen's inequality. Moreover, it can be shown that the difference between $\log p(\w_{t,d})$ and $\E_q[ \log p] - \E_q[\log q]$ is $KL( q \mid \mid p)$. Hence maximizing the bound in \eqref{eqn:lhood} is equivalent to minimizing the KL divergence between the variational and true posteriors. We denote the Evidence Lower BOund (ELBO) by $\mathcal{L}(\delta_a, \tau_{t,d}, \gamma_{t,d}, \phi_n, \lambda_k) = \E_q[ \log p] - \E_q[\log q]$. Since our objective defined in \eqref{eqn:kl} includes a regularization term, we therefore maximize a surrogate likelihood consisting of the ELBO minus the regularization term (see \citeauthor{Wainwright2007a}, 2007 for a review of penalized surrogate likelihoods). Hence our objective function $\mathcal{L}_{\rho}$ for some regularization $\rho$ is defined as:

\begin{equation}
\mathcal{L}_{\rho}(\delta_a, \tau_{t,d}, \gamma_{t,d}, \phi_n, \lambda_k) \triangleq \E_q[ \log p] - \E_q[\log q] - \rho r(q)~,
\end{equation}

\noindent where $\E_q[ \log p]$ expands for each term in the model, that is:

\begin{equation}
\begin{split}
\label{eqn:eblo_terms}
\E_q[ \log p] &= \sum_{k=1}^K \sum_{v=1}^V \E_q[\log p(\beta_{k,v} \mid \eta)] + \\
\sum_{a=1}^A &\sum_{p=1}^P \E_q[\log p(\kappa_{a,p} \mid \omega)] + \\
\sum_{t=1}^T & \Bigg[ \sum_{p=1}^P \sum_{k=1}^K \E_q[\log p(\alpha_{t,k,p} \mid \alpha_{t-1,k,p})] + \\
& \sum_{d=1}^{D_t} \bigg[ \E_q[\log p(\x_{t,d} \mid \kappa_a)] + \\
& \quad \E_q[\log p(\theta_{t,d} \mid \alpha_t \x_{t,d}, \Sigma_t)] + \\
& \quad \sum_{n=1}^{N_{d_t}} \Big[ \E_q[\log p(\z_n \mid \pi(\theta_{t,d}))] + \\
& \qquad \E_q[\log p(\w_{t,d,n} \mid \z_n, \beta)] \Big] \bigg] \Bigg]
\end{split}
\end{equation}

And, similarly $-\E_q[\log q]$ is the entropy term associated with each of the parameters. Some terms in \eqref{eqn:eblo_terms} are simple, and well-known from foundational topic models like LDA and CTM \cite{Blei2003,Lafferty2006}. For example, the topic distributions over words $\E_q[\log p(\beta_{k,v} \mid \eta)]$ term is found in LDA, and in the DAP model the distributions over personas for each author $\E_q[\log p(\kappa_{a,p} \mid \omega)]$ follows a similar structure. Similarly, the non-conjugate pairs for word-topic assignment $\E_q[\log p(\z_n \mid \pi(\theta_{t,d}))]$ has been studied in the CTM. For completeness, we show the expansion of the more unique terms $\E_q[\log p(\theta_{t,d} \mid \alpha_t \x_{t,d}, \Sigma_t)]$ and $\E_q[\log p(\alpha_{t,p} \mid \alpha_{t-1,p})]$ in the Appendix.

Expanding the objective function $\mathcal{L}_{\rho}$ according to the distribution associated with each parameter allows updates to be derived for each parameter. The parameters are optimized using a variational expectation-maximization algorithm, the details of the algorithm are given below.

\subsection{Variational E-Step}
During the E-step the model estimates variational parameters for each document and saves the sufficient statistics required to compute global parameters. The structure of the DAP model, while unique, has some components that mimic previous topic models. Specifically, the word-topic assignment parameter $\phi$ has the same update found in the CTM due to the Logistic-Normal $\gamma$ parameter. Hence $\phi$ has a closed form update: $\phi_{n,k} \propto \exp(\gamma_k) \beta_{k,v}$ \cite{Lafferty2006}.

%
%
Each author's persona is parameterized by a $\tau$. To find an update for $\tau$ we select ELBO terms featuring $\tau$, and then take the derivative with respect to each document and persona. The terms in the ELBO containing $\tau$ are:

\begin{equation*}
\begin{split}
\mathcal{L}_{[\tau]} &= \sum_{t=1}^T \sum_{d=1}^{D_t} \sum_{p=1}^P \tau_{t,d,p} \Big( \Psi(\delta_{a,p}) - \Psi(\sum_{i=1}^P \delta_{a,i}) \Big) +\\
& - \frac{1}{2} \Big( (\gamma_{t,d} - \hat{\alpha}_t \tau_{t,d})^\top \Sigma_t^{-1} (\gamma_{t,d} - \hat{\alpha}_t \tau_{t,d}) + \\
& \Tr(\Sigma_t^{-1} \diag(\tau_{t,d,p} (\hat{\alpha}_{t,p,k}^2 + \hat{\Sigma}_{t,k,k}))) \Big) +\\
& -\tau_{t,d,p} \log \tau_{t,d,p} + \\
& \lambda_{t,d} (\sum_{i=1}^P \tau_{t,d,i} - 1), \\
\end{split}
\end{equation*}

\noindent where the last term is the Lagrange constraint to ensure each vector $\tau_{t,d}$ must sum to one. Takings the derivative with respect to one specific document and persona we find that:

\begin{equation*}
\begin{split}
\frac{\partial \mathcal{L}}{\partial \tau_{t, d, p}} &= \Psi(\delta_{a,p}) - \Psi(\sum_{i=1}^P \delta_{a,i}) - \log \tau_{a,p} - 1 + \lambda +\\
\hat{\alpha}_{t, p} & \Sigma_t^{-1} (\gamma_{t,d} - \hat{\alpha}_{t, p} \tau_{t,d, p}) -\frac{1}{2} \Tr(\Sigma_t^{-1} \diag(\hat{\alpha}_{t,p}^2 + \hat{\Sigma}_{t}))
\end{split}
\end{equation*}

A closed form solution for $\tau_{t,d}$ doesn't exist. We therefore estimate $\tau_{t,d}$ using exponential gradient descent.

%
%
Since the model includes non-conjugate terms (much like DTM, CDTM, and CTM), an additional variational parameter $\zeta$ is introduced to preserve the lower-bound during the expansion of the term containing a non-conjugate pair: $\E_q[\log p(\z_n \mid \pi(\theta_{t,d}))]$. Taking the derivative of all terms containing $\zeta$ and setting it to zero yields an analogous closed form update to the one found in the CTM: $\hat{\zeta}_t = \sum_{k=1}^K \exp(\gamma_{t,d,k} + \hat{v}_{t,k}^2 / 2)$

%
%
Finally, the DAP model estimates a topic distribution for each document via the $\gamma_{t,d}$ parameter. To update $\gamma_{t,d}$ the terms in the ELBO featuring $\gamma_{t,d}$ are selected:

\begin{equation*}
\begin{split}
\mathcal{L}(\gamma_{t,d}) &= \sum_{t=1}^T \sum_{d=1}^{D_t} - \frac{1}{2} (\gamma_{t,d} - \hat{\alpha}_t \tau_{t,d})^\top \Sigma_t^{-1} (\gamma_{t,d} - \hat{\alpha}_t \tau_{t,d}) +\\
& \sum_{n=1}^{N_{d_t}} \gamma_{t,d} \phi_{n} - \zeta^{-1} (\sum_{k=1}^K \exp(\gamma_{t,d,k} + \hat{v}_t^2 / 2)) \\
\end{split}
\end{equation*}

Taking a derivative of these terms with respect to $\gamma_{t,d,k}$ yields:

\begin{equation}
\begin{split}
\label{eqn:gamma}
\frac{\partial \mathcal{L}}{\partial \gamma_{t,d,k}} =& - \Sigma_t^{-1} (\gamma_{t,d,k} - \hat{\alpha}_{t,1:P,k} \tau_{t,d,k}) + \\
& \sum_{n=1}^{N_{d_t}} \phi_{n,k} -\frac{N_{d_t}}{\zeta} \exp(\gamma_{t,d,k} + \hat{v}_{t,k}^2 / 2))\\
\end{split}
\end{equation}

Since a closed form solution isn't available, a conjugate gradient algorithm is run using the gradient in \eqref{eqn:gamma}.

%
%
Whereas $\gamma_{t,d}$ represents the mean of the Logistic-Normal for a document's topic distribution, the parameter $\hat{v}_{t,d}$ is the variance. The ELBO terms featuring $\hat{v}_{t,d}$ are:

\begin{equation*}
\begin{split}
\mathcal{L}(\hat{v}_t) &= \Tr(\Sigma_t^{-1} \hat{v}) \sum_{k=1}^K \frac{1}{2}( \log \hat{v}_k^2 + \log 2\pi) - \\
& N_{d_t}\zeta^{-1} \sum_{k=1}^K \exp(\gamma_{t,d,k} + \hat{v}_{t,k}^2 / 2) \\
\end{split}
\end{equation*}

Therefore, setting the derivative of $\mathcal{L}(\hat{v}_{t,d})$ with respect to $\hat{v}_{t,d}$ to zero and solving yields:

\begin{equation*}
\begin{split}
\frac{\partial \mathcal{L}}{\partial v_{t,d,k}^2} &= \Sigma_{t,k,k}^{-1} + \frac{1}{2\hat{v}_{t,d,k}^2} - \frac{N_{d_t}}{2\zeta} \exp(\gamma_{t,d,k} + \hat{v}_{t,k}^2 / 2),  \\
\end{split}
\end{equation*}

\noindent which requires Newton's method for each coordinate, constrained such that $\hat{v}_{t,k} > 0, \forall k$.

%
%
The parameter $\hat{\alpha}_t$ represents the noisy estimate of $\alpha_t$. After calculating $\hat{\alpha}_t$, the forward and backward equations will be applied in the M-step to give a final posterior estimate $\alpha_t$. The terms in the ELBO containing $\hat{\alpha}_t$ are found by expanding $\E_q [\log p(\alpha_{t,p}) \mid \alpha_{t-1,p})]$ for \eqref{eqn:ahat1} and $\E_q [\log p(\theta_{t,d} \mid \alpha_t \x_{t,d}, \Sigma_t)]$ for \eqref{eqn:ahat2} and \eqref{eqn:ahat3}:

\begin{subequations}
\label{eqn:ahat}
\begin{align}
\mathcal{L}(\hat{\alpha}) =& \sum_{t=1}^T \sum_{p=1}^P -\frac{1}{2} (\hat{\alpha}_{t,p} - \hat{\alpha}_{t-1,p})^\top \Sigma_t^{-1} (\hat{\alpha}_{t,p} - \hat{\alpha}_{t-1,p}) + \label{eqn:ahat1} \\
& \sum_{t=1}^T \sum_{d=1}^{D_t} -\frac{1}{2} \Big( (\gamma_{t,d} - \hat{\alpha}_t \tau_{t,d})^\top \Sigma_t^{-1} (\gamma_{t,d} - \hat{\alpha}_t \tau_{t,d}) + \label{eqn:ahat2} \\
& \sum_{t=1}^T \sum_{d=1}^{D_t} \sum_{p=1}^P \Tr \Big[ \Sigma_t^{-1} \diag \Big( \tau_{t,d,p} (\hat{\alpha}_{t,p} \hat{\alpha}_{t,p}^\top + \hat{\Sigma}_{t}) \Big) \Big]\Big) \label{eqn:ahat3}
\end{align}
\end{subequations}

Taking the derivative with respect to the mean term for each persona gives the closed form update:

\begin{equation}
\label{eqn:alpha_hat}
\hat{\alpha}_{t,p} = \frac{\hat{\alpha}_{t-1,p} + \sum_{d=1}^{D_t} (\gamma_{t,d} - 1) \tau_{t,d,p}}{1 + \sum_{d=1}^{D_t}  \tau_{t,d,p}^2}
\end{equation}

We solve for $\hat{\alpha}_{t,p}$ sequentially over time steps. For the initial time step $t=1$, we use the prior $\mu_0$ in place of $\hat{\alpha}_{t-1,p}$. Note that the summations in \eqref{eqn:alpha_hat} are collected during the E-step and $\hat{\alpha}_{t,p}$ need only be computed once after performing inference on all documents.

%
%
\subsection{Regularized Variation Inference}
\label{sec:penalty}
Our RVI algorithm nudges $\alpha_t$ to find topic distributions that are different for each persona. A natural choice for capturing this idea is an inner product between each of the personas (excluding a persona with itself). Hence, we define the regularization function by:

\begin{equation}
\label{penalty}
\rho r(q) = \sum_{p=1}^P \sum_{1 \leq q \leq P, q \not = p} \frac{D_t}{2} \rho \hat{\alpha}_{t,p}^\top \Sigma_t^{-1} \hat{\alpha}_{t,q}~,
\end{equation}

The parameter $\Sigma_t^{-1}$ in included in the regularization for two reasons. First, it simplifies the update to $\hat{\alpha}_{t,p}$. In \eqref{eqn:ahat} the term $\Sigma^{-1}$ appears in every term, which allows it to be factored out and canceled. By including $\Sigma^{-1}$ in the regularization the same cancellation can occur. Second, since $\Sigma_t^{-1} \propto I$ then its inclusion has the effect of encouraging personas to be orthogonal to one another. We include the number of documents $D_t$ at time $t$ in $r(q)$ so that the regularization is applied evenly, regardless of dataset size or a skewed distribution of documents over time. After including the regularization term in \eqref{penalty} with the ELBO terms in \eqref{eqn:ahat}, the regularized $\hat{\alpha}_{t,p}$ update is:

\begin{equation}
\label{eqn:ahat_update}
(1 + \sum_{d=1}^{D_t} \tau_{d,p}^2) \hat{\alpha}_{t,p} + \rho D_t \sum_{q \not = p} \hat{\alpha}_{t,q} = \hat{\alpha}_{t-1,p} + \sum_{d=1}^{D_t} (\gamma_d - 1) \tau_{d,p} \\
\end{equation}

Since the vector $\sum_{d=1}^{D_t} (\gamma_d - 1) \tau_{d,p}$ (of length $K$) is computed during the E-step, then the RHS is known. Similarly, the term $(1 + \sum_{d=1}^{D_t} \tau_{d,p}^2)$ is known, and in combination with $\rho D_t$ form the weights over the unknown vector $\hat{\alpha}_{t,p}$, also of length $K$. Therefore, \eqref{eqn:ahat_update} can be solved as a system of linear equations. Through experiments we've found an optimal value of $\rho \in [0, 0.5]$. The model exhibits sensitivity to the hyperparameter $\rho$, if $\rho$ is large (e.g. $>1.0$) then model quality drops due to personas overfitting to a single topic. Since $\hat{\alpha}$ is only used to estimate the global parameter $\alpha$ during the M-step, computing $\hat{\alpha}$ isn't necessary for inference on holdout datasets.

\subsection{M-Step}
In the M-step the global parameters $\alpha$, $\kappa$, and $\beta$ are updated such that the lower bound of the log likelihood of the data is maximized. Note, the update for $\beta$ is exactly the same as derived for the LDA model, and hence omitted.

%
%
The parameter $\delta$ represents the distribution over personas for each author. The closed form update for $\delta_{a,p}$ is:

\begin{equation*}
\delta_{a,p} \propto \omega_p + \sum_{t=1}^T \sum_{d=1}^{D_t} \tau_{t,d,p}
\end{equation*}

\noindent shows that $\delta$'s closed form update is an average of the persona assignments, smoothed by the author-persona prior $\omega$.

%
%
Once the variational observations $\hat{\alpha}_{t, p}$ are computed, our approach follows the variational Kalman filtering method from Wang's Continuous Time Dynamic Topic Model, see Appendix for further details. Specifically, we employ the Brownian motion to model time dynamics. However, because the DAP model's time-varying parameter is a distribution over latent topics, it performs best on data discretized in time (resulting in a smaller $T$). The forward equations mimic a Kalman filter:

\begin{equation*}
\begin{split}
m_{t,p} &= \frac{\hat{\alpha}_{t,p} P_{t,p} + m_{t-1,p} \hat{w}_t}{P_{t,p} + \hat{w}_t} \\
V_{t,p} &= \hat{w}_t \frac{P_{t,p}}{P_{t,p} + \hat{w}_t}
\end{split}
\end{equation*}

\noindent where $\hat{w}_t$ is the known process noise, and $P_{t,p} = V_{t-1,p} + \sigma \Delta_{s_t}$ captures the increase in variance as time between data points grows. Finally, the backward equations:

\begin{equation*}
\begin{split}
\alpha_{t-1,p} &= m_{t-1,p} \frac{\sigma \Delta_{s_t}}{P_{t,p}} + \alpha_{t,p} \frac{V_{t-1,p}}{P_{t,p}} \\
\Sigma_{t-1,p} &= V_{t-1,p} + \frac{(V_{t-1,p})^2}{(P_{t,p})^2} (\Sigma_{t,p} - P_{t,p}) ~, \\
\end{split}
\end{equation*}

\noindent give the updates to the remaining global parameters.

%% file: 5_experiments.tex
\section{Experiments}
\label{experiments}

\subsection{CaringBridge Dataset}
The creation of our model is inspired by a desire to discover topics on a unique dataset consisting of 14 million journals posted by half a million authors on the social networking site CaringBridge (CB). Established in 1997, CaringBridge is a 501(c)(3) non-profit organization focused on connecting people and reducing the feelings of isolation that are often associated with a patient's health journey. Due to their content, CB data has been anonymized prior to analysis.

From the CB dataset we draw an evaluation dataset consisting of journals written by authors who posted, on average, at least twice a month over a one year period. Journal posts are only kept if they contain 10 or more words. These constraints help identify a set of active users. From the 123K authors meeting these criteria, 2,000 were randomly selected. Journals written by these 2,000 authors total 114,532. Overall, authors in this dataset journal an average of 57 times, with a mean of 5 days between journal posts.

The journal texts were pre-processed in a standard way: any HTML and non-ascii (including emojis) were removed; hyphenated words and contractions were split; excess whitespace was ignored; texts were tokenized and common stopwords along with words appearing in over 90\% of documents were removed; all punctuation was stripped; and, words were reduced to their lemmas. Finally, the texts were transformed to a bag of words format, keeping the 5,000 most used words as a vocabulary set. Because the dataset includes journals written between 2006 and 2016 the timestamps are transformed into a relative value and discretized, reflecting the number of weeks since an authors first journal.

\subsection{Evaluation}
Journals are split into training and test sets with 90\%  of each author's journals ($N=103,018$) for training and 10\% ($N=11,728$) for testing. Further, variance in model performance is estimated by repeating this splitting procedure for 10-fold cross validation.

The performance of our model is compared to three other models representing the state-of-the-art in this area. The first model for comparison is LDA, which ignores authorship and temporal structure in the data. In order to evaluate LDA's performance over time, we train LDA on time steps up through $t-1$ and testing on time step $t$ (similar to the evaluation method in \citeauthor{Wang2006a}, 2006). The DTM also serves as an important baseline for comparison because it models the evolution of topics over discrete time steps. Lastly, we compare out model to CDTM, which builds on DTM and introduces a continuous treatment of time. Following the approach of others, we simply fix the number of topics at 25 for all models. The number of personas learned by the DAP model is fixed at 15.

To evaluate the models we compute the per-word log-likelihood ($PWLL$) on heldout data, which measures how well the model fits the data and is computed by $PWLL = \frac{\sum_{d=1}^D \log p(\w_d)}{\sum_{d=1}^D N_d}$. Note that perplexity, another common metric used to compare topic models, is related to $PWLL$ via $perplexity = \exp (-PWLL)$. It has been shown that perplexity (and hence $PWLL$s) don't correlate with a model finding coherent topics \cite{Chang2009}. Nevertheless, $PWLL$s provide a fair way to compare how well each model optimizes their objective functions.

%% file: 6_results.tex
\section{Results}
\label{results}
In addition evaluating model fit, we perform a qualitative analysis of the DAP model to highlight the quality and usefulness of the personas discovered. In particular, we establish that the personas are unique from one-another and capture meaningful experiences shared by authors.

\subsection{Model Comparison}
\begin{table}
\centering
\begin{tabular}{lrr}
\toprule
Model & Per-word Log-Likelihood & Std. Dev. \\
\midrule
DAP ($\rho$=0.0)  & -7.22 & 0.04 \\
DAP ($\rho$=0.2)  & -6.47 & 0.04 \\
LDA                 & -9.23  & 0.02  \\
DTM                 & -9.65  & 0.03  \\
CDTM                & -8.82 & 0.03 \\
\bottomrule
\end{tabular}
\caption{Overall comparison of models. Per-word log-likelihoods for documents in the test dataset are computed. Standard deviation in performance computed over the cross-validation sets. While the basic DAP model without regularization performs significantly better than competing model, the RVI approach further increases log-likelihoods.}
\label{overall_results}
\end{table}

In Table \ref{overall_results} we list the per-word log-likelihood and standard deviation between cross-validation sets for each of the competing models. There is a significant improvement in the DAP model's performance after regularization. Further analysis of the likelihood computation reveals that the regularization term contributes a relatively small drop in likelihood compared to the total likelihood during training. Nevertheless, these results show that even a small amount of regularization can nudge the model to seek out quality results. In testing additional $\rho$ values we found that, in general, $\rho \in [0.1, 0.3]$ faired comparably. Larger values of $\rho$ can cause model instability and the document likelihoods to have long-tailed distributions. The emergence of outlier document-likelihoods is unsurprising, regularization encourages the personas to focus on different topics --- hence, large values of $\rho$ inevitably result in personas that overfit.

\begin{figure}
\centering
\includegraphics[width=0.95\linewidth]{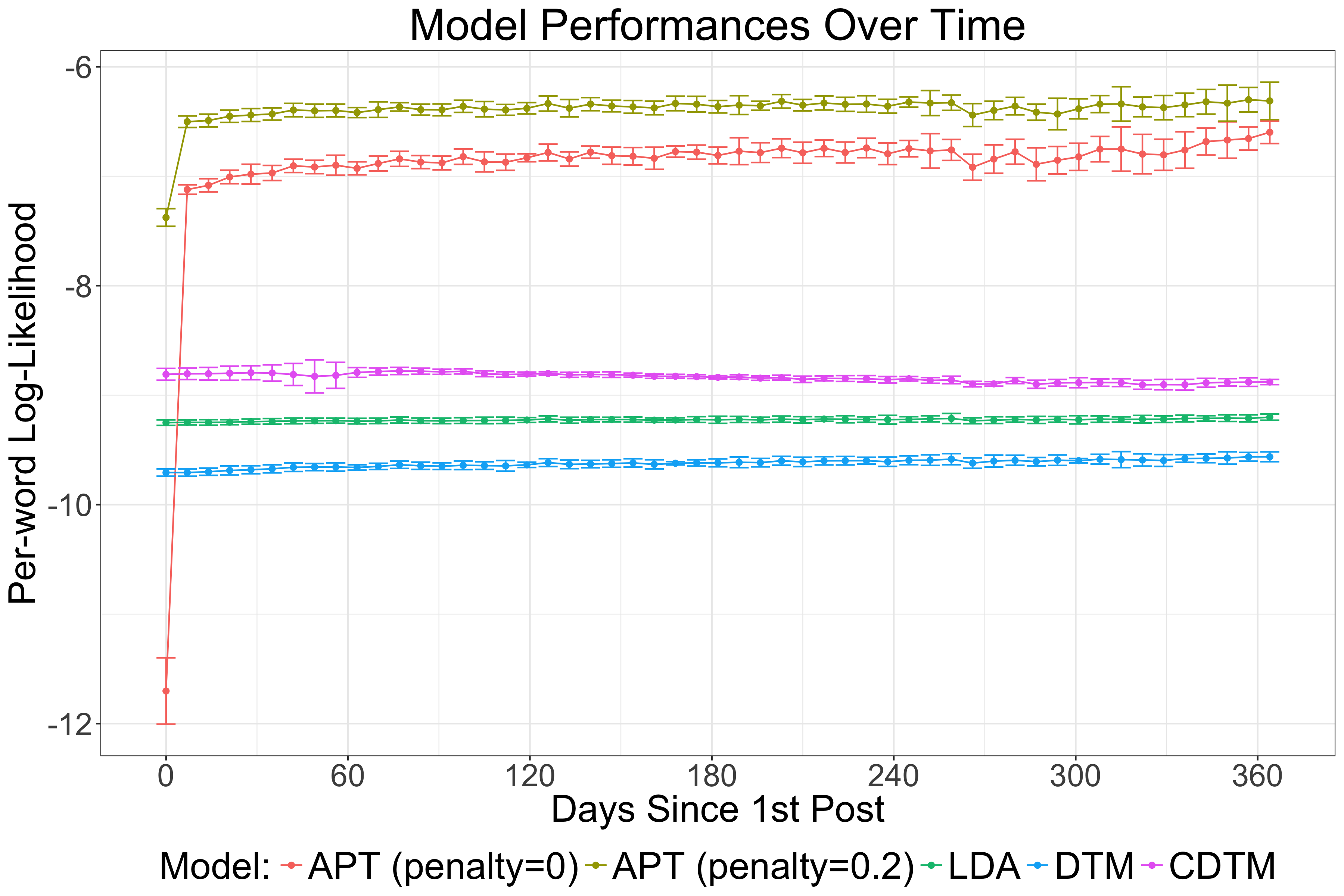}
\caption{In general the DAP model performs better than competing models over time steps. The regularized DAP model further improves performance and reduces variable results found in the first time step of the unregularized model. Error bars show one standard deviation in document-level PWLL.}
\label{fig:over_time}
\end{figure}

Figure \ref{fig:over_time} shows mean per-word log-likelihoods at each time step. The best performing DAP model shows consistently better results over competing models. However, the unregularized DAP model has a significant drop in performance in the first time step.

\subsection{Persona Quality}
\begin{figure*}
\centering
\includegraphics[width=0.9\textwidth]{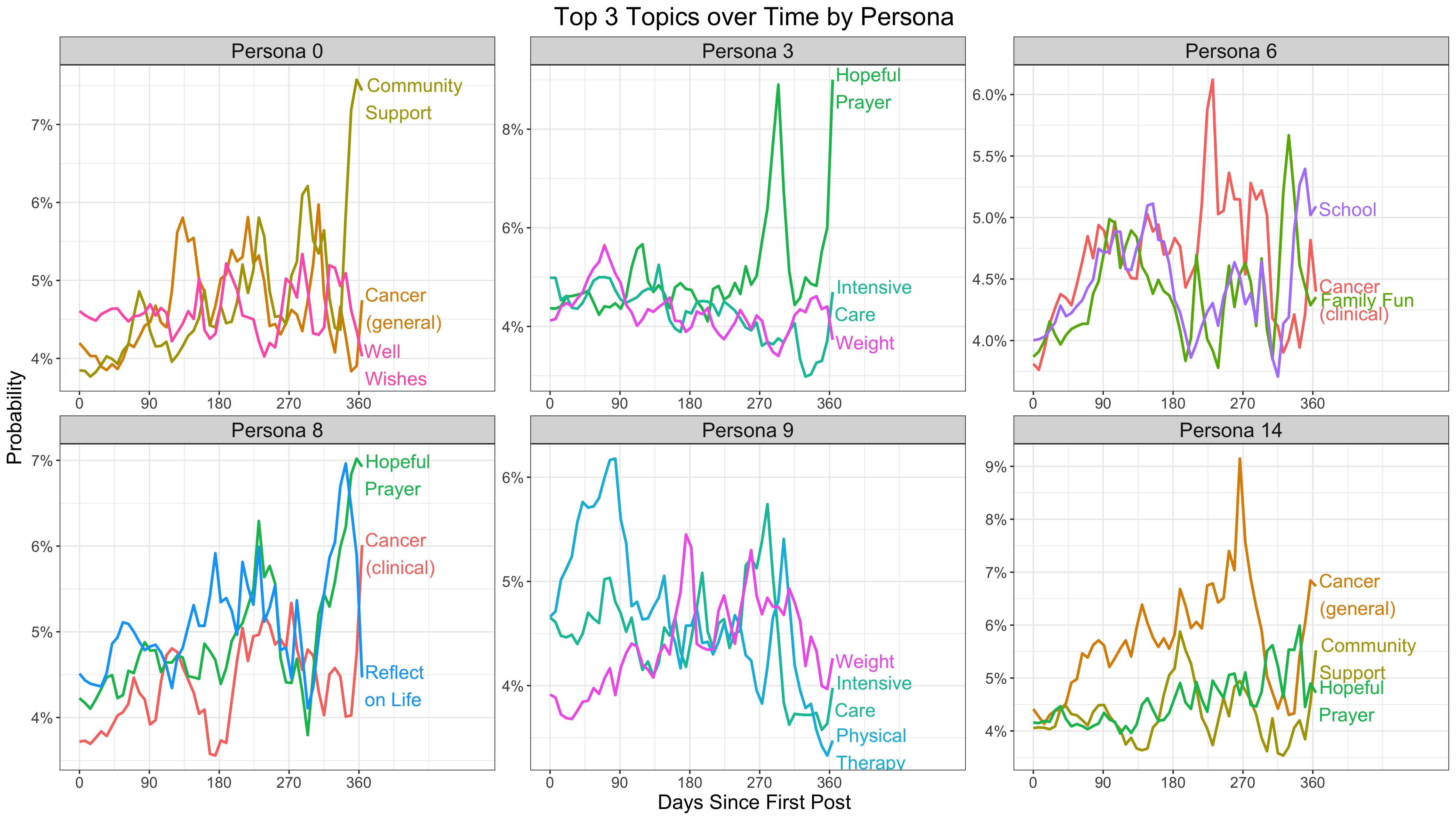}
\caption{The unregularized DAP model finds compelling, unique personas corresponding to common health journeys experienced by CaringBridge users. The three most likely topics for personas are plotted over time. Results shown for six personas that highlight diversity in topic focus. Personas 0, 6, 8, and 14 highlight nuances in how an author writes about a topic like cancer. Personas 0 and 14 engage with their community, and are less clinical when writing about cancer. Persona 14's journals, however, are more religious and often include prayer. On the other hand, when discussing health, Personas 6 and 8 write about cancer using clinical terminology. When persona 6 is not sharing health updates the conversation is often on school, family, and celebrations. Whereas, persona 8's non-health updates are deep, reflective, and prayerful.}
\label{fig:personas}
\end{figure*}

\begin{table*}
\centering
\begingroup\small
\begin{tabular}{llllllll}
\toprule
Community Support & Physical Therapy & Reflect on Life & Hopeful Prayer & Family Fun & Infection & Weather & School \\
\midrule
family & therapy & life & god & christmas & blood & nice & school \\
friend &  rehab &  know &  pray &  play &  infection &  weather &  shot \\
church &  therapist &  child &  prayer &  birthday &  fluid &  walk &  go \\
thank &  physical &  never &  lord &  game &  fever &  lunch &  appt \\
card &  pt &  love &  bless &  fun &  antibiotic &  cold &  class \\
love &  chair &  year &  please &  kid &  pressure &  snow &  tomorrow \\
service &  speech &  live &  heal &  party &  kidney &  outside &  grandma \\
friends &  progress &  people &  trust &  year &  iv &  breakfast &  teacher \\
support &  move &  cancer &  peace &  enjoy &  lung &  rain &  home \\
gift &  arm &  moment &  continue &  dinner &  clot &  go &  aunt \\
\bottomrule
 & & & & & & &  \\
\toprule
Cancer (clinical) & Cancer (general) & Intensive Care & Well Wishes & Hair Loss & Surgery & Bedtime & Weight \\
\midrule
chemo & cancer & tube & dad & hair & surgery & sleep & weight \\
blood &  treatment &  breathe &  mom &  leg &  surgeon &  night &  mommy \\
count &  radiation &  oxygen &  everyone &  wear &  heart &  bed &  gain \\
bone &  scan &  lung &  message &  head &  dr &  wake &  feed \\
marrow &  chemo &  feed &  guestbook &  look &  office &  nurse &  daddy \\
platelet &  tumor &  x\_ray &  please &  cut &  op &  say &  bottle \\
round &  oncologist &  chest &  prayer &  knee &  procedure &  asleep &  pound \\
clinic &  dr &  nurse &  read &  hat &  cardiologist &  \_time\_ &  feeding \\
transfusion &  ct &  vent &  visit &  wig &  valve &  room &  oz \\
\_url\_ &  result &  stomach &  update &  shave &  ha &  tell &  milk \\
\bottomrule
\end{tabular}
\endgroup
\caption{Top 10 words associated with the most prevalent topics found by the DAP model ($\rho=0.2$). Topic labels are selected manually in order to aid reference with Figure \ref{fig:personas}. The words \_time\_ and \_URL\_ refer to the result of text pre-processing steps for capturing common patterns like the time of day and website URLs, respectively.}
\label{tbl:words}
\end{table*}

To evaluate the quality of personas, we focus on three key elements: authors are described by one clear persona; personas are distinct from one another, as shown in the combination of topics most associated with that persona; and personas capture a coherent health journey experienced by authors.

{\bf 1:1 Author-Persona Mappings.} Authors are modeled as a distribution over personas; however, to create interpretable results we want these distributions to focus on a single persona. The DAP model achieves this in the majority of cases: 71\% of authors are concentrated on a single persona ($>90$\% probability for that persona), and 27\% of authors are evenly split between two personas. This shows that, in general, the model finds personas that generalize well enough to describe the majority of authors.

{\bf Distinct Personas.} The DAP model includes a regularization term specifically for encouraging personas with unique combinations of topics. We examined the top three topics associated with each persona. In the unregularized model, the 15 personas are only a mix of 6 different topics. In fact, a topic on "Weather" appears as a common topic for all 15 personas. On the other hand, the regularized DAP model's personas are a mix of 18 different topics. Further, the most frequently appearing topic is "Cancer (general)" (in 6 of 15 personas), which is appropriate given that approximately half of authors report cancer as a health condition.

{\bf Personas Reflect Coherent Health Journeys.} In Figure \ref{fig:personas} we show the top three topics evolving over time for selected personas. The labels listed for each topic are created manually based on words and journals most associated with the topic. Words most associated with each topic are listed in Tables \ref{tbl:words}. The persona plots in Figure \ref{fig:personas} paint a compelling picture of common health journeys experienced by CB users.

Personas reflect broad trends, often encompassing a range of health journeys. Consider Persona 9, which reflects health journeys beginning with a physical element, such as physical therapy or a health issue taking a physical toll, followed by intensive care and attention to weight. Many Persona 9 authors begin physical therapy following an accident, or are caring for a premature baby or child with a congenital disorder. However, there are a number of rare disorders that follow Persona 9's pattern. For instance, one Persona 9 author writes about a family member with Guillain-Barr\'e syndrome, a rare rapid-onset disorder in which the immune system attacks the nervous system resulting in muscle pain, weakness, and even paralysis. The syndrome often requires admittance to an intensive care unit, followed by rehabilitation -- all common themes of Persona 9.

%% file: 7_conclusion.tex
\section{Conclusion}
\label{conclusion}

The Dynamic Author-Persona topic model is uniquely suited to modeling text data with a temporal structure and written by multiple authors. Unlike previous temporal topic models, DAP discovers latent personas --- a novel component that identifies authors with similar topics trajectories. Our RVI algorithm further improves the DAP model's performance over competing models and results in the discovery of distinct personas. In evaluating the DAP model, we introduce the CaringBridge dataset: a massive collection of journals written by patients and caregivers, many of who face serious, life-threatening illnesses. From this dataset the DAP model extracts compelling descriptions of health journeys.

Many opportunities exist for further research. Currently, we deal with non-conjugate terms using the approach established in the CTM. Recent advances in non-conjugate inference \cite{Ranganath2013a,Kingma2013a,Khan2015a,Khan2017,Srivastava2017} may lead to a more efficient approach to dealing with these difficult terms.

%% file: 8_acknowledgements.tex
\section*{Acknowledgments}
We thank reviewers for their valuable comments, University of Minnesota Supercomputing Institute (MSI) for technical support, and CaringBridge for their support and collaboration. The research was supported by NSF grants IIS-1563950, IIS-1447566, IIS-1447574, IIS-1422557, CCF-1451986, CNS-1314560.

%% file: 9_appendix.tex
\newpage
\appendix
\section{ELBO Terms Unique to the DAP Model}
The expanion of the ELBO referenced in \eqref{eqn:eblo_terms} includes a number of terms previously derived for the LDA and CTM \cite{Blei2003,Lafferty2006}. The DAP model's introduction of personas, and the parameters $\z$, $\alpha$, and $\kappa$ that govern them lead to a few new terms. Terms unique to the DAP model are detailed below.

\subsection{Expanding \texorpdfstring{$\E_q[\log p(\theta_{t,d} \mid \alpha_t \x_{t,d}, \Sigma_t)]$}{Lg}}
Expansion of the ELBO term $\sum_{t=1}^T \sum_{d=1}^D \E_q[\log p(\theta_{t,d} \mid \alpha_t \x_{t,d}, \Sigma_t)]$ is unique to the DAP model, and particularly challenging because the topic distribution for each document $\theta_{t,d}$ is drawn from a Gaussian with mean $\alpha_t \x_{t,d}$. Hence, the term is expanded to:

\begin{equation*}
\begin{split}
\label{eqn:unique_expectation}
\sum_{t=1}^T & \sum_{d=1}^{D_t} \E_q[\log p(\theta_{t,d} \mid \alpha_t \x_{t,d}, \Sigma_t)] = \sum_{t=1}^T \sum_{d=1}^{D_t} \frac{1}{2} \log |\Sigma_t^{-1}| -\\
& (K/2) \log 2 \pi - \frac{1}{2} \E_q[(\theta_{t,d} - \alpha_t \x_{t,d})^\top \Sigma_t^{-1} (\theta_{t,d} - \alpha_t \x_{t,d})]
\end{split}
\end{equation*}

Note that the expectation in $\E_q[(\theta_{t,d} - \alpha_t \x_{t,d})^\top \Sigma_t^{-1} (\theta_{t,d} - \alpha_t \x_{t,d})]$ is over all the terms --- that is, $\alpha_t \x_{t,d}$ are not constants. Factorizing this expectation gives:

\begin{subequations}
\label{eqn:theta_factors}
\begin{align}
\E_q & [(\theta_{t,d} - \alpha_t \x_{t,d})^\top \Sigma_t^{-1} (\theta_{t,d} - \alpha_t \x_{t,d})] = \nonumber \\
& \E_q[\theta_{t,d}^\top \Sigma^{-1}_t \theta_{t,d}] \label{eqn:theta_factors1} +\\
& - 2 \E_q[\theta_{t,d}^\top \Sigma_t^{-1} \alpha_t \x_{t,d}] \label{eqn:theta_factors2} + \\
& \E_q[(\alpha_t \x_{t,d})^\top \Sigma_t^{-1} (\alpha_t \x_{t,d})] \label{eqn:theta_factors3} ~,
\end{align}
\end{subequations}

\noindent where each of the terms in \eqref{eqn:theta_factors} is evaluated below.

{\bf Term \eqref{eqn:theta_factors1}:} Since $\E_q[\theta_{t,d}^\top \Sigma^{-1}_t \theta_{t,d}]$ is a straight-forward case of the Guassian quadratic identity:

\begin{equation*}
\E_q[\theta_{t,d}^\top \Sigma^{-1}_t \theta_{t,d}] = \Tr(\Sigma_t^{-1} \hat{v}_{t,d}) + \gamma_{t,d} \Sigma_t^{-1} \gamma_{t,d}~,
\end{equation*}

\noindent where $\gamma_{t,d}$ is the variational parameter for $\theta_{t,d}$ and $\hat{v}_{t,t}$ is the variance parameter associated with the topic distribution over document $d_t$.

{\bf Term \eqref{eqn:theta_factors2}:} Doesn't take a Guassian quadratic form. To solve $- 2 \E_q[\theta_{t,d}^\top \Sigma_t^{-1} \alpha_t \x_{t,d}]$, recall that  $\theta_{t,d}$ and $\alpha_t \x_{t,d}$ are independent under the the mean-field assumption, thus:

\begin{equation*}
\begin{split}
- 2 \E_q[\theta_{t,d}^\top \Sigma_t^{-1} \alpha_t \x_{t,d}] &= -2 \Big( \E_q[\theta_{t,d}] \Sigma_t^{-1} \E_q[\alpha_t \x_{t,d}] \Big) \\
&= -2(\gamma_{t,d} \Sigma_t^{-1} \hat{\alpha}_t \tau_{t,d}) \\
\end{split}
\end{equation*}

{\bf Term \eqref{eqn:theta_factors3}:} Expanding the last term yields:

\begin{equation*}
\begin{split}
\E_q[(\alpha_t \x_{t,d})^\top \Sigma_t^{-1} (\alpha_t \x_{t,d})] &= \E_q[(\alpha_t \x_{t,d})] \Sigma_t^{-1} \E_q[(\alpha_t \x_{t,d})] \\
&= \hat{\alpha}_t \tau_{t,d} \Sigma_t^{-1} \hat{\alpha}_t \tau_{t,d} + \Tr(\Sigma_t^{-1} S)~,
\end{split}
\end{equation*}

\noindent where $S = \E_q[(\alpha_t \x_{t,d})(\alpha_t \x_{t,d})^\top]$ denotes the variance of the $\alpha_t \x_{t,d}$ terms. To evaluate $S$, consider the variance between personas $i$ and $j$, which simplifies the computation because $x_i$ and $x_j$ are scalars and $\alpha_{t,i}$ refers to a column vector of $\alpha_t$:

\begin{equation*}
\begin{split}
\E_q[(\alpha_{t,i} x_{t,d,i})(\alpha_{t,j} x_{t,d,j}^\top)] &= \E_q[\alpha_{t,i} \alpha_{t,j}^\top   x_{t,d,i} x_{t,d,j}] \\
&= \E_q[\alpha_{t,i} \alpha_{t,j}^\top] \E_q[x_{t,d,i} x_{t,d,j}] \\
\end{split}
\end{equation*}

The resulting $K \times K$  covariance matrix has off-diagonal elements are all 0 since $\x$ -- a draw from a multinomial -- is a 1 of $P$ vector. Elements along the diagonal are given by $\E_q[\alpha_{t,i} \alpha_{t,j}^\top] = \hat{\alpha}_{t,i} \hat{\alpha}_{t,i}^\top + \hat{\Sigma}_{t}$. Thus, for persona $p$ we have $S = \diag( \tau_{t,d,p} (\hat{\alpha}_{t,p} \hat{\alpha}_{t,p}^\top + \hat{\Sigma}_{t}))$. Therefore term \eqref{eqn:theta_factors3} is:

\begin{equation*}
\hat{\alpha}_t \tau_{t,d} \Sigma_t^{-1} \hat{\alpha}_t \tau_{t,d} + \sum_{p=1}^P \Tr \Big[ \Sigma_t^{-1} \diag \Big( \tau_{t,d,p} (\hat{\alpha}_{t,p} \hat{\alpha}_{t,p}^\top + \hat{\Sigma}_{t}) \Big) \Big] \\
\end{equation*}

Combining the three expanded terms from \eqref{eqn:theta_factors} can be reduced:

\begin{equation*}
\begin{split}
& \Tr(\Sigma_t^{-1} \hat{v}_t) + \gamma_{t,d} \Sigma_t^{-1} \gamma_{t,d} - 2(\gamma_{t,d} \Sigma_t^{-1} \hat{\alpha}_t \tau_{t,d}) + \\
& \quad \hat{\alpha}_t \tau_{t,d} \Sigma_t^{-1} \hat{\alpha}_t \tau_{t,d} + \sum_{p=1}^P \Tr \Big[ \Sigma_t^{-1} \diag \Big( \tau_{t,d,p} (\hat{\alpha}_{t,p} \hat{\alpha}_{t,p}^\top + \hat{\Sigma}_{t}) \Big) \Big] \\
&= (\gamma_{t,d} - \hat{\alpha}_t \tau_{t,d})^\top \Sigma_t^{-1} (\gamma_{t,d} - \hat{\alpha}_t \tau_{t,d}) + \\
& \quad \Tr(\Sigma_t^{-1} \hat{v}_t) + \sum_{p=1}^P \Tr \Big[ \Sigma_t^{-1} \diag \Big( \tau_{t,d,p} (\hat{\alpha}_{t,p} \hat{\alpha}_{t,p}^\top + \hat{\Sigma}_{t}) \Big) \Big]
\end{split}
\end{equation*}

Finally, the ELBO term for $\E_q[\log p(\theta_{t,d} \mid \alpha_t \x_{t,d}, \Sigma_t)]$ is expanded out fully to:

\begin{equation*}
\begin{split}
\sum_{t=1}^T & \sum_{d=1}^{D_t} \E_q[\log p(\theta_{t,d} \mid \alpha_t \x_{t,d}, \Sigma_t)] = \\
& \sum_{t=1}^T \sum_{d=1}^{D_t} \frac{1}{2} \log |\Sigma_t^{-1}| - (K/2) \log 2 \pi - \\
& \frac{1}{2} \Big( (\gamma_{t,d} - \hat{\alpha}_{t} \tau_{t,d})^\top \Sigma_t^{-1} (\gamma_{t,d} - \hat{\alpha}_{t} \tau_{t,d}) + \Tr(\Sigma_t^{-1} \hat{v}_t) + \\
& \quad \sum_{p=1}^P \Tr \Big[ \Sigma_t^{-1} \diag \Big( \tau_{t,d,p} (\hat{\alpha}_{t,p} \hat{\alpha}_{t,p}^\top + \hat{\Sigma}_{t}) \Big) \Big] \Big) +\\
\end{split}
\end{equation*}

\subsection{Expanding \texorpdfstring{$\E_q[\log p(\alpha_{t,p} \mid \alpha_{t-1,p})]$}{Lg}}
Expanding the ELBO term $\E_q[\log p(\alpha_{t,p} \mid \alpha_{t-1,p})]$ is similar to the DTM, and follows from the Gaussian quadratic form identity, which states that: $\E_{m,V} (x - \mu)^\top \Sigma^{-1} (x - \mu) = (m - \mu)^\top \Sigma^{-1} (m - \mu) + \Tr(\Sigma^{-1} V)$.

\begin{equation*}
\begin{split}
\sum_{t=1}^T & \sum_{p=1}^P \E_q [ \log p(\alpha_{t,p} \mid \alpha_{t-1,p})] = \\
& \sum_{t=1}^T \sum_{p=1}^P \frac{1}{2} \log |\Sigma_t^{-1}| - \frac{K}{2} \log 2 \pi - \\
& \quad \frac{1}{2} \E_q[(\alpha_{t,p} - \alpha_{t-1,p})^\top \Sigma_t^{-1} (\alpha_{t,p} - \alpha_{t-1,p})] \\
&= \sum_{t=1}^T \sum_{p=1}^P \frac{1}{2} \log |\Sigma_t^{-1}| - \frac{K}{2} \log 2 \pi - \\
& \quad \frac{1}{2} \Big( (\hat{\alpha}_{t,p} - \hat{\alpha}_{t-1,p})^\top \Sigma_t^{-1} (\hat{\alpha}_{t,p} - \hat{\alpha}_{t-1,p}) + \Tr(\hat{\Sigma}_t \Sigma_t^{-1})  \Big) \\
\end{split}
\end{equation*}

\section{Time Dynamics}
The time dynamics of the DAP model follow the variational Kalman filtering from \citeauthor{Wang2008}. Below we derive the forward and backward equations for updating $\alpha$.

To clarify notation the parameters $\alpha$ and $\hat{\alpha}$ are defined in the more conventional notation associated with Kalman Filters. The updates are defined for each persona $p$ indicating the need to estimate the the distribution over topics for each persona $p$ (for each $\alpha_t$ and $\hat{\alpha}_t$).

\begin{equation*}
\begin{split}
\alpha_{t,p} &= \alpha_{t-1,p} + w_t \\
w_t &\sim \mathcal{N}(0, \sigma \Delta_{s_t}),
\end{split}
\end{equation*}

\noindent where $\Delta_{s_t}$ denotes the different in time between time stamps $s_t$ and $s_{t-1}$, and $\sigma$ is the "process noise" which is implemented as a constant factor. Similarly, the variational parameters are defined by:

\begin{equation*}
\begin{split}
\hat{\alpha}_{t,p} &= \alpha_{t,p} + v_t \\
v_t &\sim \mathcal{N}(0, \hat{v}_t),
\end{split}
\end{equation*}

Note, $\hat{v}_t$ is the "measurement noise" which is also implemented as a constant factor. Since the dynamics we employ follow the Brownian motion model, the Kalman Filters state transition matrix $\Phi_t = I$.

\paragraph{Forward Equations}
Define $m_{t,p} = \E[ \alpha_{t,p} \mid \hat{\alpha}_{i \leq t, p}]$ and $V_{t,p} = \E[ (\alpha_{t,p} - m_{t,p})^2 \mid \hat{\alpha}_{i \leq t, p}]$. Then the forward equations when a persona is not observed are:

\begin{equation*}
\begin{split}
m_{t,p} &= m_{t-1,p} \\
V_{t,p} &= P_{t,p} \\
P_{t,p} &= V_{t-1,p} + \sigma \Delta_{s_t},
\end{split}
\end{equation*}

In other words, when no new information is observed for a persona, the prior for the forward mean becomes the posterior, and the variance grows proportional to the difference in time elapsed. Since personas are latent, it's unlikely that a persona will be completely unobserved in a timestep. In practice we implement a threshold checking if $\sum_{t=1}^T \sum_{d=1}^{D_t} \tau_{t,d,p} > 1$ to determine if persona $p$ is observed.

On the other hand, when enough authors exhibiting some persona are observed, the measurement residual, residual covariance, and Kalman gain are:

\begin{equation*}
\begin{split}
r_t &= \hat{\alpha}_{t,p} - m_{t-1,p} \\
S_t &= P_{t,p} + \hat{v}_t \\
K_t &= P_{t,p} (S_t)^{-1} = \frac{P_{t,p}}{P_{t,p} + \hat{v}_t} \\
\end{split}
\end{equation*}

Thus, the forward equation for the mean is then:

\begin{equation*}
\begin{split}
m_{t,p} &= m_{t-1,p} + K_t r_t \\
&= m_{t-1,p} + \frac{P_{t,p}}{P_{t,p} + \hat{v}_t} (\hat{\alpha}_{t,p} - m_{t-1,p}) \\
&= \hat{\alpha}_{t,p} \frac{P_{t,p}}{P_{t,p} + \hat{v}_t} + m_{t-1,p} \Big( 1 - \frac{P_{t,p}}{P_{t,p} + \hat{v}_t} \Big) \\
&= \hat{\alpha}_{t,p} \frac{P_{t,p}}{P_{t,p} + \hat{v}_t} + m_{t-1,p} \frac{\hat{v}_t}{P_{t,p} + \hat{v}_t}\\
&= \frac{\hat{\alpha}_{t,p} P_{t,p} + m_{t-1,p} \hat{v}_t}{P_{t,p} + \hat{v}_t}
\end{split}
\end{equation*}

The forward equation for the variance is then:

\begin{equation*}
\begin{split}
V_{t,p} &= (I - K_t) P_{t,p} \\
&= (I - \frac{P_{t,p}}{P_{t,p} + \hat{v}_t}) P_{t,p} \\
&= \frac{\hat{v}_t}{P_{t,p} + \hat{v}_t} P_{t,p} \\
&= \hat{v}_t \frac{P_{t,p}}{P_{t,p} + \hat{v}_t}
\end{split}
\end{equation*}

\paragraph{Backward Equations}
For the backward equations first define:

\begin{equation*}
\begin{split}
\alpha_{t,p} \mid \hat{\alpha}_{i : i \leq T, p} &\sim \mathcal{N}(\widetilde{m}_{t,p}, \widetilde{V}_{t,p}) \\
\widetilde{m}_{t,p} &= \E [ \alpha_{t,p} \mid \hat{\alpha}_{i : i\leq T, p} ] \\
\widetilde{V}_{t,p} &= \E [ (\alpha_{t,p} - \widetilde{m}_{t,p})^2 \mid \hat{\alpha}_{i : i \leq T, p} ], \\
\end{split}
\end{equation*}

\noindent hence

\begin{equation*}
\begin{split}
\widetilde{m}_{t,p} &= m_{t,p} + \frac{V_{t,p}}{P_{t+1,p}} (\widetilde{m}_{t+1,p} - m_{t+1,p}) \\
\widetilde{V}_{t,p} &= V_{t,p} + \frac{V_{t,p}}{P_{t+1,p}} (\widetilde{V}_{t,p} - P_{t,p}) \frac{V_{t,p}}{P_{t+1,p}} \\
\end{split}
\end{equation*}

The backwards equations for documents observed one time index after the other are:

\begin{equation*}
\begin{split}
\widetilde{m}_{t-1,p} &= m_{t-1,p} + \frac{V_{t-1,p}}{P_{t,p}} (\widetilde{m}_{t,p} - m_{t,p}) \\
&= m_{t-1,p} \Big( 1 -  \frac{V_{t-1,p}}{P_{t,p}} \Big) + \widetilde{m}_{t,p} \frac{V_{t-1,p}}{P_{t,p}} \\
&= m_{t-1,p} \Big( 1 -  \frac{V_{t-1,p}}{V_{t-1,p} + \sigma \Delta_{s_t}} \Big) + \widetilde{m}_{t,p} \frac{V_{t-1,p}}{P_{t,p}} \\
&= m_{t-1,p} \frac{\sigma \Delta_{s_t}}{P_{t,p}} + \widetilde{m}_{t,p} \frac{V_{t-1,p}}{P_{t,p}} \\
\widetilde{V}_{t-1,p} &= V_{t-1,p} + \frac{V_{t-1,p}}{P_{t,p}} (\widetilde{V}_{t,p} - P_{t,p}) \frac{V_{t-1,p}}{P_{t,p}} \\
&= V_{t-1,p} + \frac{(V_{t-1,p})^2}{(P_{t,p})^2} (\widetilde{V}_{t,p} - P_{t,p}) \\
\end{split}
\end{equation*}